\PassOptionsToPackage{main=english,russian}{babel}
\PassOptionsToPackage{T2A}{fontenc}
\documentclass[onecolumn]{misraj}

\usepackage{lipsum}
\usepackage{pdfpages}
\usepackage{colortbl}
\usepackage{tabulary}
\usepackage{longtable}
\usepackage{array}
\usepackage{placeins}
\usepackage{tablefootnote}
\usepackage{tcolorbox}
\usepackage{wrapfig}
\usepackage{adjustbox}
\usepackage{float}
\usepackage{caption}
\usepackage{breqn} 

\usepackage{pifont}
\usepackage{tabularx}

\usepackage{arydshln}

\usepackage{xspace} %
\usepackage{makecell} 
\usepackage{multirow}

\usepackage[framemethod=TikZ]{mdframed}
\usepackage{multicol}

\usepackage[utf8]{inputenc}
\usepackage[T1]{fontenc}
\usepackage[english]{babel} 
\usepackage{arabtex}
\usepackage{utf8}
\setcode{utf8}

\definecolor{lightblue}{RGB}{211, 227, 252} 
\definecolor{bgblue}{RGB}{247, 250, 255} 

\newcommand*\colourcheck[1]{%
  \expandafter\newcommand\csname #1check\endcsname{\textcolor{#1}{\ding{52}}}%
}
\newcommand*\colourcross[1]{%
  \expandafter\newcommand\csname #1cross\endcsname{\textcolor{#1}{\ding{55}}}%
}
\DeclareSymbolFont{extraup}{U}{zavm}{m}{n}
\DeclareMathSymbol{\vardiamond}{\mathalpha}{extraup}{87}
\colourcheck{blue}
\colourcheck{green}
\colourcheck{red}

\colourcross{blue}
\colourcross{green}
\colourcross{red}
\usepackage{nicematrix}
\usepackage{comment}
\usepackage{amssymb}
\usepackage[bottom]{footmisc}
\usepackage{svg}
\usepackage{spverbatim}
\RequirePackage[hidelinks,colorlinks=true,linkcolor=blue,citecolor=blue]{hyperref}
\setcitestyle{number,square}
\definecolor{deeppurple}{HTML}{9e02f7}
\definecolor{forestgreen}{HTML}{2e7d43}

\hypersetup{urlcolor=deeppurple}

\usepackage{booktabs}
\usepackage{amsmath}
\usepackage{pgfplots}
\pgfplotsset{compat=1.18}
\usetikzlibrary{positioning,arrows.meta,fit,backgrounds,shapes.geometric,calc}
\usepackage{enumitem}
\setlist{itemsep=2pt,topsep=4pt}

\definecolor{gainc}{HTML}{2e7d43}
\definecolor{lossc}{HTML}{c0392b}
\definecolor{basec}{HTML}{9aa5b1}
\definecolor{smallc}{HTML}{5b8def}
\definecolor{largec}{HTML}{1f3f8f}
\definecolor{boxframe}{HTML}{3b5bdb}

\newcommand{\task}[1]{\textsc{#1}}
\newcommand{\WSP}{WSP\xspace}
\newcommand{\bud}[1]{b#1}

\newtcolorbox{finding}[1][]{colback=bgblue,colframe=boxframe,boxrule=0.6pt,arc=2pt,
  left=6pt,right=6pt,top=4pt,bottom=4pt,fonttitle=\bfseries\small,title=#1}

\title{Edits Without Abilities: Weight-Space Programming on a Small Vision-Language Model}

\multiauthors
\author{
    name={Khalil Hennara},
    email={hennara@misraj.ai}
}
\author{
    name={Muhammad Hreden},
    email={hreden@gmail.com}
}

\author{
    name={Sara Chrouf},
    email={sara.chrouf@misraj.ai}
}
\author{
    name={Safwan AlModhayan},
    email={safwan@misraj.ai}
}

\affiliations{
 \includegraphics[width=2cm]{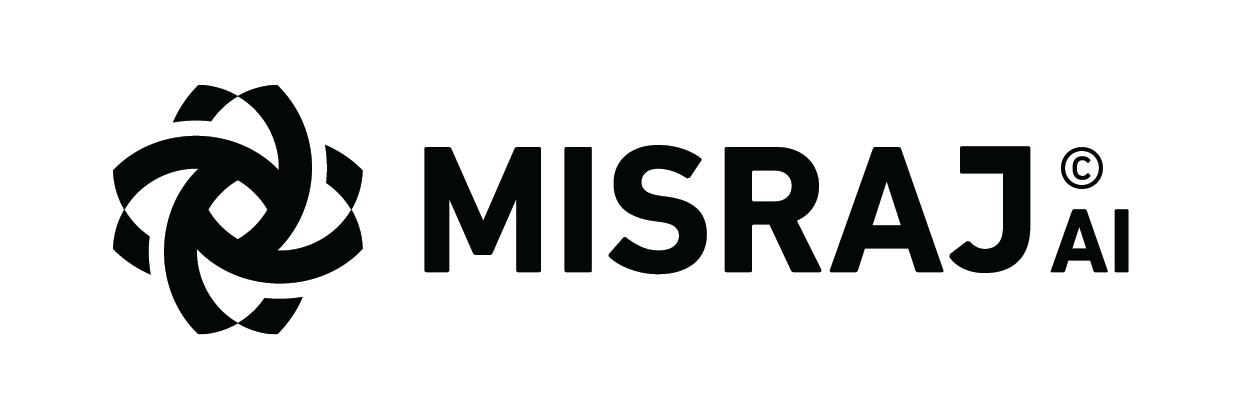}\\
     Khobar, Saudi Arabia \\
     {hennara,sara.chrouf,safwan}@misraj.ai \\
}
\date{\today}
\abstract{
Weight-space programming (\WSP) asks a frontier language model to improve another network by editing its weights directly, with no gradients and no training loop. Its headline demonstration raised a small language model's math score several-fold, but the gain came mostly from removing reasoning loops. We test what the idea can and cannot do on a harder target: a vision-language model on six document and reasoning tasks (Arabic and English handwriting, Arabic and English document OCR, Arabic table extraction, and visual math). An agent edits the weights under a fixed budget of forward passes. Across twelve runs, the agent found real, closed-form edits: \textit{last-layer readout corrections, position-keyed rank-1 updates and hand-built gated neurons}. On the private test sets they cut character error rate by up to 46\% on English OCR and 20\% on Arabic handwriting. The edits changed how the model writes, not what it reads: where the model lacked the underlying skill, as in reading Arabic handwriting, no edit supplied it. Where the scoring setup rewarded a proxy, as in math under a 256-token limit, the agent optimised the proxy, and the edited model did worse on the private set. We also find that suppressing a token hands its probability to the next wrong alternative, that teacher-forced checks overstate edit quality, and that the agent acts as an iterative optimiser whose development gains shrink on held-out data. Agents also tended to commit to one or two families of edits early. We release the harness, the data splits and the full agent logs, and we argue that \WSP needs a human in the loop to steer the objective and the search.
}

\begin{document}

\renewcommand{\thefootnote}{\fnsymbol{footnote}}

\section{Introduction}
\label{sec:introduction}

Can a language model improve another neural network by reasoning about its weights? \citet{takehi2026sparks} call this \emph{weight-space programming}: a general-purpose model reads a network's weights and writes changes into them directly, without an optimiser and without gradient descent. Their most cited result is an edit to Qwen3.5-0.8B that raised MATH-500 accuracy from 7.9\% to 48.1\% by changing about 0.016\% of the weights. Critics soon pointed out that the gain came almost entirely from removing a pathology. In the base model, 93\% of answers ran into the 32{,}768-token limit because of reasoning loops, and the edit let them finish \citep{honnibal2026misleading,barry2026lp,wang2026closer}. The original authors' own control made the same point: on Gemma~3 270M, which rarely loops, edits barely moved the score. They concluded that weight editing could remove an obstacle hiding an ability but could not add an ability the model lacked \citep{takehi2026sparks}.

That leaves the interesting question open. When the target has no single switch to flip, what does weight-space programming actually change? We study this on a target with a much larger and more varied weight space: a small \emph{vision-language} model, with a vision tower, a connector and a language model. We pick tasks where the model is known to struggle but where the failures are of different kinds: reading Arabic, which the model barely knows; long-form document OCR, where it often loops or chooses the wrong output format; table extraction, where a wrong header ruins the rest; and visual math. Each task has a private test set kept on a separate machine, so the agent can never see what it will be graded on.

Our goal is not to reproduce a headline number, but to find where the method stops working and why. We therefore read the logs of all twelve agent runs (two budgets per task) turn by turn and report what the agent changed, what it believed, and what held up on the private sets.

\paragraph{Findings in brief.}
\begin{itemize}
  \item \textbf{The levers are formatting, not perception}: edits changed the output mode, annotation conventions and verbosity, rarely what the model reads (\S\ref{sec:levers}).
  \item \textbf{No new knowledge}: where reading was the bottleneck (Arabic handwriting), no edit supplied it (\S\ref{sec:knowledge}).
  \item \textbf{The agent optimises the proxy}: on math it shortened replies to fit a 256-token limit, and private accuracy fell (\S\ref{sec:proxy}).
  \item \textbf{Recurring mechanisms}: suppressing a token promotes the next wrong one, and teacher-forced checks overstate edit quality (\S\ref{sec:mechanisms}).
  \item \textbf{The agent is an optimiser whose search narrows early}, which motivates a human in the loop (\S\ref{sec:agent}, \S\ref{sec:hitl}).
\end{itemize}

\paragraph{Contributions.}
\begin{itemize}
    \item An open harness\footnote{\url{https://github.com/misraj-ai/wsp-vlm}} for weight-space programming on Hugging Face vision-language models. It separates the agent machine from the evaluation machine, freezes every edit into a verifiable bundle\footnote{\url{https://huggingface.co/datasets/Misraj/wsp_bundle/tree/main}}, and enforces a forward-pass budget and a no-gradient policy. 
    \item Twelve agent runs with full logs\footnote{\url{https://huggingface.co/datasets/Misraj/wsp_runs/tree/main}} on six tasks, each graded on a private test set\footnote{\url{https://huggingface.co/datasets/Misraj/wsp_vl_test/tree/main}}, and a new Arabic financial-table extraction dataset. 
\end{itemize}
\section{Background and Related Work}
\label{sec:background}

\paragraph{Weight-space programming.} \citet{takehi2026sparks} introduced the term for a general-purpose LLM that reads and writes a network's weights by reasoning, without an optimiser. Their headline edit raised Qwen3.5-0.8B on MATH-500 from 7.9\% to 48.1\%. Follow-up analyses traced this gain to the removal of reasoning loops under a pathological baseline \citep{honnibal2026misleading}. They showed that the edit is a form of activation steering \citep{wang2026closer} and that a linear-program solver finds a comparable edit with far fewer changed weights \citep{barry2026lp}. Replications with other frontier models reproduce the same loop fix \citep{kundu2026samefix,goworek2026replicating}.

\paragraph{Model editing and steering.} Established ways to change a model's behaviour directly include rank-one updates to MLP output matrices with covariance-based keys, as in ROME and MEMIT \citep{meng2022rome,meng2023memit}; activation steering with difference-of-means directions \citep{turner2023actadd,rimsky2024caa}; and directional ablation \citep{arditi2024refusal}. This literature reports two limits. First, steering directions pick up spurious features of their contrast data and generalise unreliably \citep{tan2024steering}, and prompting or fine-tuning often beats them \citep{wu2025axbench}. Second, in multimodal models, edits on the language side work far better than edits on the vision side \citep{cheng2023mmedit,huang2024vlkeb}.

\paragraph{Optimising a proxy.} An optimiser that maximises a measurable target can satisfy its letter and miss its intent \citep{krakovna2020specification,manheim2018goodhart}, and frontier agents have been observed doing this in research-like tasks \citep{metr2025rewardhacking}. For this reason, systems built for agentic research keep a scientist in the loop to set objectives and steer the search \citep{gottweis2025coscientist}.

\paragraph{Where our study fits.} Earlier weight-space programming studies used language-only models and public benchmarks, and their headline result rested on a single pathology. We test the method where neither holds: a vision-language model on six tasks with different kinds of failure. We propose no new editing method and no competing training baseline; this is an analysis study. We ask which known edit families the agent rediscovers, whether the limits reported for steering and multimodal editing also bind an agent that edits weights freely (they do), and how the agent, acting as an optimiser of the practice score, comes to optimise proxies without any cheating.

\section{Method}
\label{sec:method}

\subsection{Problem setup}

The agent receives a target model, a task description, a practice set split into development and holdout halves, and a budget of $B$ forward passes. It may read the weights and run arbitrary Python in a kernel, but it may not compute gradients, use optimisers, import model-running libraries, or run the model itself. The only way to run the model is the harness's forward-pass tool, and each call costs one pass. Edits are written through a validated store that refuses changes to the token embeddings, the output head and the final norm, as well as non-finite values, shape changes and whole-tensor rescaling. The run ends with a submission: a set of changed tensors plus a short explanation. Following the original protocol, the agent must use its whole budget; an attempt to stop early is refused.

\subsection{Harness}

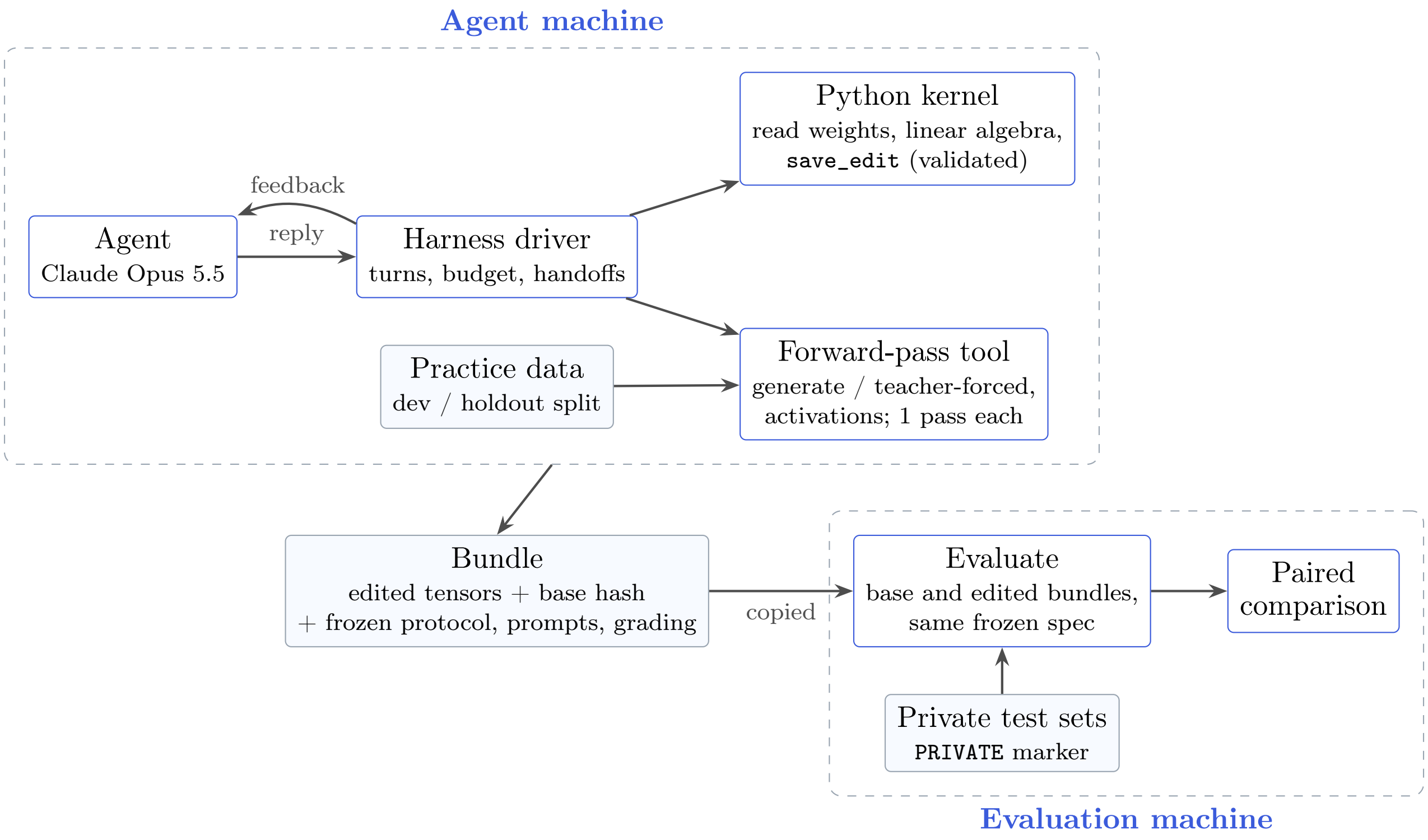
\begin{figure}[t]
\centering
\begin{tikzpicture}[
  box/.style={draw=boxframe,rounded corners=2pt,fill=white,align=center,font=\small,inner sep=4pt,minimum height=0.9cm},
  dat/.style={draw=basec,rounded corners=2pt,fill=bgblue,align=center,font=\small,inner sep=4pt,minimum height=0.9cm},
  arr/.style={-{Stealth[length=2.2mm]},thick,draw=black!70},
  lab/.style={font=\scriptsize,align=center,text=black!70}]
  \node[box] (agent) {Agent\\[-1pt]\scriptsize Claude Opus 5.5};
  \node[box,right=1.4cm of agent] (driver) {Harness driver\\[-1pt]\scriptsize turns, budget, handoffs};
  \node[box,above right=0.35cm and 1.2cm of driver] (kernel) {Python kernel\\[-1pt]\scriptsize read weights, linear algebra,\\[-2pt]\scriptsize \texttt{save\_edit} (validated)};
  \node[box,below right=0.35cm and 1.2cm of driver] (fwd) {Forward-pass tool\\[-1pt]\scriptsize generate / teacher-forced,\\[-2pt]\scriptsize activations; 1 pass each};
  \node[dat,below=0.55cm of driver] (practice) {Practice data\\[-1pt]\scriptsize dev / holdout split};
  \draw[arr] (agent) -- node[above,lab]{reply} (driver);
  \draw[arr] (driver.north west) ++(0,-0.1) to[bend right=25] node[above,lab,pos=0.5]{feedback} (agent.north east);
  \draw[arr] (driver) -- (kernel);
  \draw[arr] (driver) -- (fwd);
  \draw[arr] (practice) -- (fwd);
  \node[draw=basec,dashed,rounded corners=4pt,fit=(agent)(driver)(kernel)(fwd)(practice),inner sep=8pt,
        label={[font=\small\bfseries,text=boxframe]above:Agent machine}] (am) {};
  \node[dat,below=1.25cm of practice] (bundle) {Bundle\\[-1pt]\scriptsize edited tensors + base hash\\[-2pt]\scriptsize + frozen protocol, prompts, grading};
  \draw[arr] (am.south) ++(0,0) -- (bundle.north);
  \node[box,right=1.7cm of bundle] (eval) {Evaluate\\[-1pt]\scriptsize base and edited bundles,\\[-2pt]\scriptsize same frozen spec};
  \node[dat,below=0.55cm of eval] (private) {Private test sets\\[-1pt]\scriptsize \texttt{PRIVATE} marker};
  \node[box,right=0.9cm of eval] (cmp) {Paired\\[-1pt]comparison};
  \draw[arr] (bundle) -- node[below,lab]{copied} (eval);
  \draw[arr] (private) -- (eval);
  \draw[arr] (eval) -- (cmp);
  \node[draw=basec,dashed,rounded corners=4pt,fit=(eval)(private)(cmp),inner sep=8pt,
        label={[font=\small\bfseries,text=boxframe]below:Evaluation machine}] {};
\end{tikzpicture}
\caption{The harness. The agent edits weights on one machine with practice data only. Each submission is frozen with the protocol into a bundle and graded on a different machine that holds the private test sets.}
\label{fig:harness}
\end{figure}

Figure~\ref{fig:harness} shows the two-machine design. The agent machine holds the base weights, the practice data and the agent loop. The private test sets never touch it. At the end of a run, the submission is frozen into a \emph{bundle}: the edited tensors, a hash of the base weights, and the full specification of the model, decoding protocol, task prompts and grading. Bundles are copied to the evaluation machine, which refuses any folder without a \texttt{PRIVATE} marker and evaluates the base bundle and every edited bundle under the same frozen specification. Comparisons are paired by item and refuse results produced under different prompts, decoding or grading.

\paragraph{The forward-pass tool.} A pass either generates outputs for up to $N$ practice items, with the original model, the edited model, or both on the same items, or reads given target texts with teacher forcing. Teacher-forced passes can return log-probabilities and activations from chosen layers. Generation results are graded with the task metric and summarised with paired statistics, loop and cut-off rates. Failed requests are not charged. Between turns, long sessions are compressed into a handoff that keeps the agent's own notebook.

\paragraph{The kernel.} The agent's Python runs in a persistent kernel with the weights mounted read-only. A static policy refuses autograd, optimisers, model libraries, subprocesses and network access. Calls to linear solvers trigger an advisory warning only: closed-form solves are allowed, and they are the agent's main tool. Our runs used Google Colab, where the bubblewrap sandbox cannot start, so the kernel ran unsandboxed (\texttt{sandbox: none}). This did not expose the private data, which was never on the agent machine. It did let one agent read the Hugging Face cache;
\begin{finding}[The Agent reading cache `Arabic OCR task`]
There is a tokenizer.json in the local Hugging Face cache, next to the weights.

tokj = json.load(open(/root/.cache/huggingface/hub/../tokenizer.json))

\end{finding}

\subsection{Tasks, data and model}

All experiments use Qwen3.5-0.8B, a 24-layer early-fusion vision-language model with a 1024-wide language model and a tied output head \citep{qwen2026qwen35}. We study six tasks (Table~\ref{tab:data}). Every task has a practice split, which the agent uses through the forward-pass tool, and a private test split, which exists only on the evaluation machine. All data available at Hugging Face\footnote{\url{https://huggingface.co/collections/Misraj/wsp}}
\begin{itemize}
  \item \task{ar-hw}: the Arabic handwritten-line dataset of the NAKBA-NLP 2026 shared task, which we used in earlier work \citep{hennara2026nakba}. Practice items come from its training split and test items from its test split.
  \item \task{en-hw}: the line-level release of the IAM handwriting database by Teklia \citep{teklia2024iamline,marti2002iam}, with the same use of its training and test splits.
  \item \task{ar-ocr}: tested on 400 pages of Misraj-DocOCR \citep{hennara2025baseer}. The 1{,}000 practice \footnote{\url{https://huggingface.co/datasets/Misraj/Misraj-DocOCR_Train_Split}} pages come from the same distribution as the data used to train Baseer, and we release them with this paper.
  \item \task{table} \footnote{\url{https://huggingface.co/datasets/Misraj/Financial-data_Train_Split}}: a new Arabic table-extraction dataset built for this study.
  \item \task{en-ocr}: OmniDocBench \citep{ouyang2025omnidocbench}, used for full-page extraction.
  \item \task{math}: practice on DynaMath, whose questions are generated by programs \citep{zou2025dynamath}; test on 246 questions of MATH-Vision testmini \citep{wang2024mathvision}.
\end{itemize}
Practice scores are $1-\mathrm{CER}$ for the transcription tasks, with Arabic normalisation (diacritics, tatweel and letter variants) for the Arabic tasks. The table score is $0.5\times$ header-cell F1 $+\,0.5\times$ body-cell F1 on the first table in the output; on the private set all non-math tasks, including \task{table}, are reported as WER and CER. Math uses answer accuracy on the content of the last \verb|\boxed{}|, with numeric tolerance.

\begin{table}[t]
\centering
\small
\caption{Data. Practice items are visible to the agent through forward passes; test items exist only on the evaluation machine.}
\label{tab:data}
\begin{adjustbox}{max width=\linewidth}
\begin{tabular}{@{}lllrrl@{}}
\toprule
Task & Content & Source & Practice & Test & Practice score \\
\midrule
\task{ar-hw} & Arabic handwritten lines & NAKBA-NLP 2026 \citep{hennara2026nakba} & 1000 & 2095 & $1-\mathrm{CER}$ \\
\task{en-hw} & English handwritten lines & IAM-line \citep{teklia2024iamline} & 1000 & 2915 & $1-\mathrm{CER}$ \\
\task{ar-ocr} & Arabic document pages & Misraj-DocOCR \citep{hennara2025baseer} & 1000 & 400 & $1-\mathrm{CER}$ \\
\task{en-ocr} & English document pages & OmniDocBench \citep{ouyang2025omnidocbench} & 500 & 250 & $1-\mathrm{CER}$ \\
\task{table} & Arabic financial documents & New (this work) & 1000 & 200 & header/body cell F1 \\
\task{math} & Visual math questions & DynaMath / MATH-Vision & 5010 &  246 & answer accuracy \\
\bottomrule
\end{tabular}
\end{adjustbox}

\vspace{2pt}
\end{table}

\subsection{Agent and runs}

The agent is Claude Opus~5.5. We chose it over the GPT-6 Astra agent used by \citet{takehi2026sparks} on the basis of a replication. We re-ran their policy-fixing experiment, in which the agent must repair a broken car-control policy network by editing its weights, under the same conditions. Opus~5.5 repaired the policy by changing 387 parameters; Astra's reported fix changed about 6{,}000. Both fixes solve the task; one that is more than ten times smaller is more targeted and easier to inspect. Together with an independent replication of the original Qwen3.5-0.8B result with Opus~5.5 \citep{kundu2026samefix}, we take this as evidence that Opus~5.5 is a suitable agent for weight-space programming. The full replication, including the logs, prompt, and videos, is available at this \href{https://drive.google.com/drive/folders/1eh-aLk40kWuHYgIXImi5Ga_MlJrlSvR_?usp=drive_link}{link}.

Each task has a single-task experiment: the agent sees practice data and feedback for that task only and is told that the model will also be tested on other tasks. We ran two budgets per task, twelve runs in total, with the settings in Table~\ref{tab:config}. A run lasted between 43 and 345 agent turns and between 31 and 226 Python executions.

\begin{table}[t]
\centering
\small
\caption{Run configuration per task. Pass size is the maximum number of items per generation pass (per model in paired passes). ``Sampled'' means temperature 0.7, top-$p$ 0.80, top-$k$ 65; ``greedy'' means no sampling.}
\label{tab:config}
\begin{tabular}{@{}llrrll@{}}
\toprule
Task & Decoding & Max tokens & Pass size & Budgets (passes) & Reasoning effort \\
\midrule
\task{ar-hw}  & sampled & 256   & 256 (128) & 10, 25 & default (10), \texttt{xhigh} (25) \\
\task{en-hw}  & greedy  & 256   & 128 (64)  & 10, 25 & default \\
\task{ar-ocr} & sampled & 2{,}048 & 16 (8)  & 25, 50 & \texttt{xhigh} \\
\task{en-ocr} & sampled & 2{,}048 & 16 (8)  & 25, 50 & \texttt{xhigh} \\
\task{table}  & sampled & 2{,}048 & 16 (8)  & 10, 30 & \texttt{xhigh} \\
\task{math}   & greedy  & 256   & 64        & 10, 30 & \texttt{xhigh} \\
\bottomrule
\end{tabular}
\end{table}

\subsection{Evaluation}

Each bundle, and the base model, is evaluated once on the private test set of its task under the frozen protocol. We report mean word and character error rates (lower is better) for the transcription and table tasks, and accuracy for math. For the analysis we use the agents' own practice-holdout measurements and every turn of their logs, which we cite as task and turn (for example \task{en-ocr} t072).

\section{Experiments and Results}
\label{sec:experiments}

\subsection{Private test sets}

\begin{figure}[t]
\centering
\begin{tikzpicture}
\begin{axis}[
  xbar, width=0.8\linewidth, height=7.2cm,
  bar width=7pt, enlarge y limits=0.1,
  xmin=-50, xmax=15,
  xlabel={Relative change in private-set CER (\%) \qquad $\leftarrow$ better \quad worse $\rightarrow$},
  symbolic y coords={TABLE, AR-OCR B, AR-OCR A, EN-OCR, EN-HW, AR-HW},
  ytick=data, yticklabel style={font=\small\scshape},
  xtick={-50,-40,-30,-20,-10,0,10},
  xmajorgrids, grid style={black!10},
  legend style={at={(0.5,-0.2)},anchor=north,legend columns=2,font=\small,draw=none,/tikz/every even column/.append style={column sep=12pt}},
  reverse legend,
  axis line style={black!40}, tick style={black!40},
  extra x ticks={0}, extra x tick labels={}, extra x tick style={grid style={black!70,thick}},
]
\addplot[fill=largec,draw=none] coordinates {
  (9.8,AR-HW) (-12.5,EN-HW) (-46.3,EN-OCR) (-6.3,AR-OCR A) (-9.2,AR-OCR B) (5.0,TABLE)};
\addplot[fill=smallc,draw=none] coordinates {
  (-19.5,AR-HW) (-12.5,EN-HW) (-37.3,EN-OCR) (-22.2,AR-OCR A) (5.4,AR-OCR B) (0,TABLE)};
\legend{Large budget, Small budget}
\end{axis}
\end{tikzpicture}
\caption{Relative change in private-set CER against the base model. The small-budget \task{table} run submitted an empty edit (0\%). The \task{en-hw} value comes from CER rounded to two decimals, so its size is uncertain.}
\label{fig:private}
\end{figure}
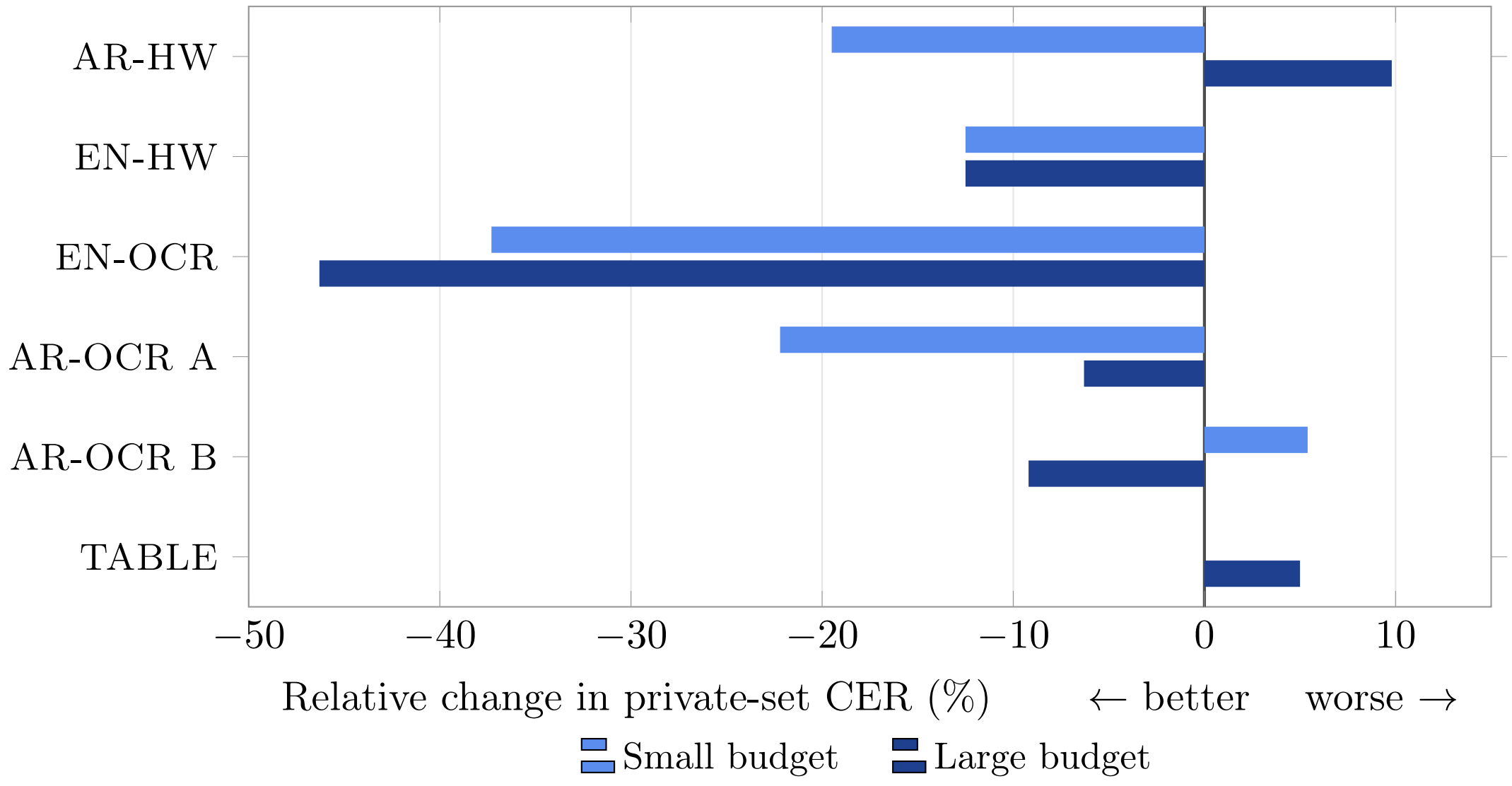

Table~\ref{tab:private} and Figure~\ref{fig:private} show the results. The largest gain is on English OCR, where CER fell from 0.67 to 0.36. Arabic handwriting improved at the small budget (CER 0.82 to 0.66), and Arabic OCR gave mixed results. On English handwriting, WER fell by 40\% while CER barely moved (\S\ref{sec:levers}). Table extraction got slightly worse, and math got clearly worse: MathVision accuracy fell from 24.8\% to 19.9\% (\bud{10}) and 17.1\% (\bud{30}) (\S\ref{sec:proxy}). A larger budget did not reliably help.

\begin{table}[t]
\centering
\small
\caption{Private test sets. WER and CER: lower is better; relative change in parentheses. Mean CER is not clipped, so values above 1 come from runaway outputs. Arabic OCR sets A and B are defined in \S\ref{sec:method}. Single evaluations; differences of a few percent are within noise.}
\label{tab:private}
\begin{adjustbox}{max width=\linewidth}
\begin{tabular}{@{}lccc@{\hspace{14pt}}ccc@{}}
\toprule
 & \multicolumn{3}{c}{WER} & \multicolumn{3}{c}{CER} \\
\cmidrule(lr){2-4}\cmidrule(lr){5-7}
Task & Base & Small budget & Large budget & Base & Small budget & Large budget \\
\midrule
\task{ar-hw} (\bud{10}/\bud{25})   & 1.12 & \textbf{0.99} \scriptsize($-$12\%) & 1.04 \scriptsize($-$7\%) & 0.82 & \textbf{0.66} \scriptsize($-$20\%) & 0.90 \scriptsize(+10\%) \\
\task{en-hw} (\bud{10}/\bud{25})   & 0.35 & 0.21 \scriptsize($-$40\%) & \textbf{0.20} \scriptsize($-$43\%) & 0.08 & 0.07 & 0.07 \\
\task{en-ocr} (\bud{25}/\bud{50})  & 0.77 & 0.61 \scriptsize($-$21\%) & \textbf{0.50} \scriptsize($-$35\%) & 0.67 & 0.42 \scriptsize($-$37\%) & \textbf{0.36} \scriptsize($-$46\%) \\
\task{ar-ocr}, set A (\bud{25}/\bud{50}) & 1.45 & 1.38 \scriptsize($-$5\%) & \textbf{1.37} \scriptsize($-$6\%) & 1.58 & \textbf{1.23} \scriptsize($-$22\%) & 1.48 \scriptsize($-$6\%) \\
\task{ar-ocr}, set B (\bud{25}/\bud{50}) & 1.20 & 1.52 \scriptsize(+27\%) & \textbf{1.13} \scriptsize($-$6\%) & 1.30 & 1.37 \scriptsize(+5\%) & \textbf{1.18} \scriptsize($-$9\%) \\
\task{table} (\bud{10}/\bud{30})   & 1.65 & = base\textsuperscript{a} & 1.66 \scriptsize(+1\%) & 1.39 & = base\textsuperscript{a} & 1.46 \scriptsize(+5\%) \\
\midrule
 & \multicolumn{6}{c}{MathVision accuracy, \% (higher is better; 246 questions)} \\
 & \multicolumn{2}{c}{Base} & \multicolumn{2}{c}{Small budget (\bud{10})} & \multicolumn{2}{c}{Large budget (\bud{30})} \\
\task{math} & \multicolumn{2}{c}{\textbf{24.8}} & \multicolumn{2}{c}{19.9 \scriptsize($-$4.9)} & \multicolumn{2}{c}{17.1 \scriptsize($-$7.7)} \\
\task{ar-hw} \bud{10} edit on math & \multicolumn{2}{c}{24.8} & \multicolumn{2}{c}{\textbf{27.6} \scriptsize(+2.8)} & \multicolumn{2}{c}{--} \\
\bottomrule
\end{tabular}
\end{adjustbox}

\vspace{2pt}
{\footnotesize \textsuperscript{a}The \bud{10} run submitted an empty edit, so its outputs are those of the base model.}
\end{table}

\subsection{What the agent built}

Table~\ref{tab:summary} summarises each task. On five of six tasks, the score was limited by how the model writes (format or convention), not by what it reads. The final edits are small: one or two rank-1 terms or neurons, mostly in the last MLP layer, where the effect of an edit on the next token can be computed exactly.

\begin{table}[t]
\centering
\small
\caption{Practice-side summary of the larger-budget run per task. Held-out gain is the agent's own paired estimate on practice holdout items (standard error in parentheses).}
\label{tab:summary}
\begin{tabularx}{\linewidth}{@{}lcXXc@{}}
\toprule
Task & Base & What limited the score & Final edit & Held-out gain \\
\midrule
\task{ar-hw} & 0.40 & Reading the script & Last-layer readout correction & $\approx$+0.012 \\
\task{en-hw} & 0.935 & IAM puts a space before punctuation & Rank-1 term, layer 23 & +0.013 (0.002) \\
\task{ar-ocr} & 0.13 & Writes web-page HTML instead of markdown & Gated neurons, layer 23 & +0.049 (0.020) \\
\task{en-ocr} & 0.39 & First token chooses HTML mode & Rank-1 terms, layer 23 & +0.16 (0.03) \\
\task{math} & 0.26--0.29 & 68\% of replies cut at 256 tokens & Gated ``concise'' steering, layer 14 & +0.043 (0.025) \\
\task{table} & 0.05--0.08 & Wrong layout mode; reversed Arabic & Rank-1 term, layer 23 & $-$0.039 \\
\bottomrule
\end{tabularx}
\end{table}

\paragraph{Arabic handwriting.} Wrong outputs start with one unlikely token and then lock in. Suppressing those tokens only pushed the model into other bad modes. The edit that worked was a last-layer readout correction. Its key fix came from a tokenizer detail: Arabic-Indic digits are two byte tokens (t077). On the private set, the small run beat the large one (CER 0.66 against 0.90).

\paragraph{English handwriting.} After one pass, the agent saw that IAM references put a space before punctuation (``record , anyway .''), while the model writes normal prose. Both budgets found the same rank-1 edit that adds these spaces. This explains the private result: WER falls by 40\% because each spaced mark counts as a word, while CER barely moves.

\paragraph{OCR.} In both OCR tasks, the first answer token decides the format: 75--90\% of its probability goes to \verb|<html| or a code fence. The model reads well once it is in the right mode ($-0.4$ to $-0.8$ nats per reference token). The English OCR edit calibrated the first token. The Arabic OCR edit used a gated neuron that switches prose pages to markdown and leaves table pages alone.

\paragraph{Tables.} The base model writes a web-page layout no reference uses. Both budgets tried the same format-forcing edits in the same order, and all failed. At 10 passes the agent submitted no edit. At 30 it submitted one whose blind holdout result was negative.

\section{Analysis}
\label{sec:evaluation}

\subsection{The levers are formatting, not perception}
\label{sec:levers}

Most of the gap between the model and the references was about \emph{how} to write. The agents checked this cheaply before editing. On English handwriting, a regex fix of the spacing raised the score by 0.021, and the edit reached about 60\% of that. On Arabic OCR, a rule-based markdown conversion gave +0.085, and the edits reached +0.03 to +0.05. The format choice also barely depends on the image. On \task{table}, the hidden states at the first answer token are almost identical across pages (mean cosine 0.93). The page information is there, though: on Arabic OCR, a linear readout separates table pages from prose pages with AUC 0.985. The default decision just ignores it. Format also affects reading direction. In HTML mode the model often writes Arabic reversed, for example \<ةكرش> instead of \<شركة>, and these outputs score 0.028 against 0.130 for the rest.

\begin{finding}[Takeaway 1]
Closed-form edits easily change how a small VLM writes, and hardly change what it reads. Gains often measure agreement with the scorer's conventions, and should be reported that way. Where a rule-based fix existed, it matched or beat the edit on practice data, so \WSP is most useful as a diagnostic that finds such conventions.
\end{finding}

\subsection{Edits do not add knowledge the model lacks}
\label{sec:knowledge}

Arabic handwriting is the test case. The model reads this script poorly (baseline 0.40). The agent calculated that even a perfect boost of the model's top choice would raise the chance of the right token only from 0.269 to 0.30. The one edit on the vision side, to the connector, hurt on every difficulty level. The edit that worked makes the model write what it already half-reads more consistently; it does not fix the misreading. No edit to the vision tower or connector survived in any run. This matches the original authors' finding that editing can remove an obstacle but cannot add an ability \citep{takehi2026sparks}. Gradient training on the same kind of data does teach small VLMs to read Arabic \citep{hennara2025baseer,hennara2026nakba}.

\subsection{The agent optimises the proxy it is given}
\label{sec:proxy}

On math, 67--69\% of the base model's replies hit the 256-token limit. Finished replies were right 67--75\% of the time, cut ones 4--7\%. Both runs saw this in their first pass, so the practice score became mostly a score for finishing on time, and the agents worked on that. They built a ``be concise'' steering edit from verbose replies and short rewrites. The first version also learned to skip the reasoning and cut the score from 0.266 to 0.141. The agent fixed the contrast data and gated the edit, and the final version gained +0.043 on the practice holdout.

On MathVision, both edits made the model worse: 24.8\% to 19.9\% (\bud{10}) and 17.1\% (\bud{30}). Shorter replies cost the reasoning that harder questions need. By contrast, the \bud{10} Arabic handwriting edit, made with the same budget and never shown a math question, scored 27.6\%. That is 7 more correct answers out of 246, within noise. Because that edit faces the same change of benchmark (DynaMath to MATH-Vision), neither editing itself nor the shift explains the drop; the losses come from what the math edits targeted.

\begin{finding}[Takeaway 2]
The agent optimises what the protocol measures. Here that was reply length, not math. Before trusting a gain, check what the score actually rewards.
\end{finding}

\subsection{Recurring mechanisms}
\label{sec:mechanisms}

\begin{figure}[t]
\centering
\begin{tikzpicture}[
  st/.style={draw=boxframe,rounded corners=2pt,fill=white,font=\small\ttfamily,align=center,inner sep=3pt,minimum height=0.75cm,text width=2.25cm},
  ef/.style={font=\scriptsize,align=center,text=lossc,text width=2.3cm},
  arr/.style={-{Stealth[length=2mm]},thick,draw=black!60}]
  \node[st] (a) {```, <html};
  \node[st,right=0.55cm of a] (b) {<};
  \node[st,right=0.55cm of b] (c) {<table};
  \node[st,right=0.55cm of c] (d) {<img};
  \node[st,right=0.55cm of d] (e) {"The ..."};
  \draw[arr] (a) -- (b); \draw[arr] (b) -- (c); \draw[arr] (c) -- (d); \draw[arr] (d) -- (e);
  \node[ef,below=0.15cm of a] {web-page HTML mode};
  \node[ef,below=0.15cm of b] {copies the prompt example};
  \node[ef,below=0.15cm of c] {top opener on 55 of 91 pages};
  \node[ef,below=0.15cm of d] {46\% of opening mass; loops};
  \node[ef,below=0.15cm of e] {chatty openings};
  \node[font=\small,above=0.15cm of c] {suppress the top opener $\rightarrow$ the next one takes over};
\end{tikzpicture}
\caption{Substitution chain on English OCR. Each box is the most likely opening after the previous one was suppressed; red text is the resulting failure.}
\label{fig:balloon}
\end{figure}
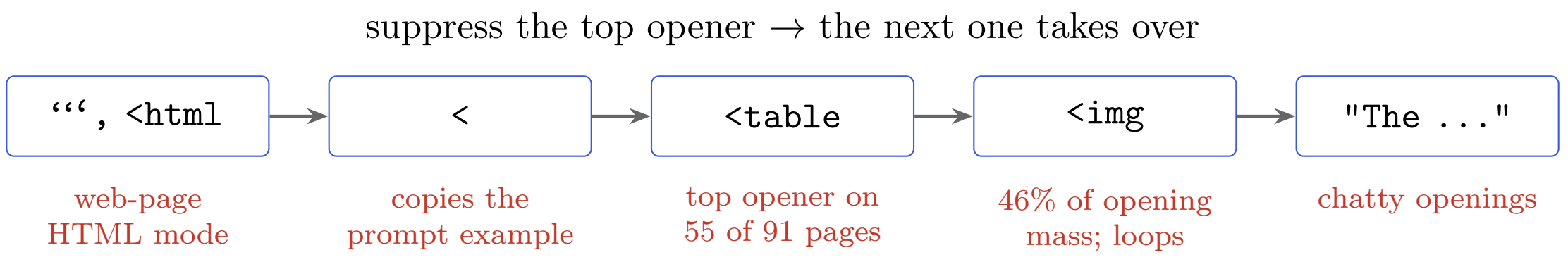

\paragraph{Suppressing a token moves its probability to a neighbour.} In five of six tasks, removing a bad token promoted the next wrong one (Figure~\ref{fig:balloon}). Softmax gives the freed probability to the runner-up, and tokens that share embedding directions move together. On \task{table}, demoting about 70 markup tokens raised the space token by 17 logits. What worked instead was calibration: matching the first-token distribution to how references begin.

\paragraph{Teacher-forced checks overstate edit quality.} Most runs built an exact one-step simulator of last-layer edits. It cannot see what happens after the first changed token. On Arabic handwriting, an edit that helped in teacher forcing lost 0.012 in generation. This is exposure bias \citep{bengio2015scheduled}. The partial fix was to check the edit on the edited model's own outputs.

\paragraph{Where an edit fires matters more than what it pushes.} On English OCR, a global bias changed the top token at about 1\% of positions and lost. The same push applied only at the first answer token changed 0.05\% and won. On Arabic OCR, a gated neuron could express ``first token \emph{and} prose page'', which a linear edit could not.

\subsection{The agent's behaviour}
\label{sec:agent}

\paragraph{The agent is the optimiser.} Each edit was closed-form, but the loop of trying, checking and patching is fitting by another name. English OCR went through 12 versions, and Arabic handwriting tried about 50 settings. As a result, the gains the agent measured on its development items shrank on held-out practice items. Here, gain is the edited model's practice score minus the original's ($1-\mathrm{CER}$ for OCR, accuracy for math). Arabic OCR fell from +0.114 to almost zero, English OCR from +0.25 to +0.09, and math from +0.078 to +0.043. Only English handwriting kept its gain (+0.016 to +0.013).

\paragraph{Holdout discipline broke where evidence was thin.} With greedy decoding and 64 pairs per pass, the standard error was 0.002. With sampling and 8 pairs, it was 0.03--0.13. All holdout misuse we found was in the 8-pair tasks, for example reusing holdout errors to design the next version.

\paragraph{More budget mostly bought confirmation.} Larger runs spent 27--68\% of their passes on holdout checks, and tried few new ideas. On English handwriting, the edit was fixed by pass 6. Budget helped only where there were many separate failures to fix, as on English OCR. Most budget pairs also differ in effort or data split, so these are not clean comparisons.

\section{Discussion}
\label{sec:discussion}

\subsection{The case for a human in the loop}
\label{sec:hitl}

Two of our findings point the same way. First, the agent optimises whatever the protocol measures, and nothing inside the loop questions whether that is the intended target (\S\ref{sec:proxy}). Second, the search narrows early. Once an agent had a plausible family of edits, it spent the rest of its budget refining and confirming it. Both math runs chose the same steering method independently; both table runs tried the same three format-forcing edits in the same order; the larger Arabic handwriting run tried about four families in 25 passes. A larger budget rarely produced a new idea. These are exactly the points where a researcher adds the most value at little cost. We therefore propose three checkpoints:
\begin{enumerate}
  \item \textbf{Approve the diagnosis and the objective.} After the first passes, the agent states what limits the score and what it intends to change. A human checks that this is the intended target, for example ``finish within the limit'' versus ``reason better''.
  \item \textbf{Choose among families.} Before committing, the agent proposes two or three qualitatively different edit families with cheap upper bounds, and a human picks or redirects.
  \item \textbf{Guard the holdout.} A human freezes the edit and runs the confirmation, so the agent cannot develop on holdout items.
\end{enumerate}
We did not run this protocol; testing it is the natural next experiment.

\subsection{Recommendations for \WSP harnesses}

The failures that recurred across tasks suggest concrete changes:
\begin{itemize}
  \item \textbf{Report power, not only pass counts.} Give each task a pass size that yields a useful standard error, and consider a separate holdout budget.
  \item \textbf{Hold other variables fixed across budget pairs}: reasoning effort, split seed and practice-data version.
  \item \textbf{Let forward requests reference saved files.} Pasting requests inline limited every math contrast set to 15--48 pairs and caused most request errors.
  \item \textbf{Provide decode-only tokenizer access.} The single most important fix in Arabic handwriting came from a tokenizer detail the agent had to reconstruct.
  \item \textbf{Keep experiment parameters in handoff notes}, not only conclusions.
  \item \textbf{Give practice data for at least one other task.} Every ``other tasks should not get worse'' check was a proxy, and one such proxy removed the useful part of an edit (\task{ar-hw} t103--t120).
  \item \textbf{Clean the practice data and check the graders.} 110 table references were leaked model reasoning text, and the math grader mis-parsed multi-value and date answers.
\end{itemize}

\subsection{Limitations}

We study one model and one agent, with two runs per task, so we cannot measure run-to-run variance. Only the English handwriting pair differs in budget alone; the other pairs also differ in reasoning effort, data split or practice data. Effects on other tasks were measured once, for the Arabic handwriting edit on math. Most private sets were evaluated once under sampling, and mean CER is not clipped, so a few runaway outputs can dominate it. We compare against no prompting or fine-tuning baseline on the private sets; rule-based fixes were measured on practice data only. Finally, the agent's kernel ran without a sandbox on Colab; the private data was safe on a separate machine.

\section{Conclusion}
\label{sec:conclusion}

Weight-space programming works on a small vision-language model, in a specific sense. A frontier agent, with no gradients, found compact edits that fixed what the model writes at decisive tokens: the output format, an annotation convention, the length of a reply. On English OCR this cut private-set character errors almost in half. It did not teach the model to read what it could not read, and where the protocol rewarded a proxy, it optimised the proxy, at the cost of real accuracy on math. The agent's own process mirrors these limits: it fits iteratively in all but name, it trusts single-step checks that miss multi-step failures, and it commits early to a narrow search. We think the most useful framing of \WSP is not as a replacement for training but as a fast, inspectable way to diagnose and patch behaviour, best run with a human who sets the objective and steers the search. We release the harness, data splits and logs to make these limits easy to test and, we hope, to move. Our analysis covers the main patterns but not every turn of the twelve runs, so we encourage readers to explore the released logs: they likely hold findings we missed.

\bibliography{main}

\end{document}